\documentclass[10pt,journal,compsoc,twoside]{IEEEtran}
\ifCLASSOPTIONcompsoc
  \usepackage[nocompress]{cite}
\else
  \usepackage{cite}
\fi
\usepackage{array}
\usepackage{graphicx}
\graphicspath{{figures/}}
\usepackage{amsmath}
\usepackage{dblfloatfix}
\usepackage{url}
\usepackage{orcidlink}

\usepackage{comment}

\title{EEG-D\textsuperscript{3}: A Solution to the Hidden Overfitting Problem of Deep Learning Models}

\author{Siegfried~Ludwig~\orcidlink{0000-0001-9693-9482},
        Stylianos~Bakas~\orcidlink{0000-0003-1054-0169},
        Konstantinos~Barmpas~\orcidlink{0000-0001-6724-3689},
        Georgios~Zoumpourlis~\orcidlink{0000-0002-5579-6953},\\
        Dimitrios~A.~Adamos~\orcidlink{0000-0001-6700-1057},
        Nikolaos~Laskaris~\orcidlink{0000-0002-1960-394X},
        Yannis~Panagakis~\orcidlink{0000-0003-0153-5210},
        Stefanos~Zafeiriou~\orcidlink{0000-0002-5222-1740}
\IEEEcompsocitemizethanks{
    \IEEEcompsocthanksitem S.~Ludwig and S.~Bakas (equal contribution), K.~Barmpas, G.~Zoumpourlis, D.~A.~Adamos and S.~Zafeiriou are with the Department
    of Computing, Imperial College London, London SW7 2RH, U.K.\protect\\
    E-mail: siegfried.ludwig20@imperial.ac.uk
    \IEEEcompsocthanksitem S. Bakas and N.~Laskaris are with the School of Informatics, Aristotle University of Thessaloniki, Thessaloniki 54124, Greece.\protect\\
    E-mail: simpakas@csd.auth.gr
    \IEEEcompsocthanksitem Y.~Panagakis is with the Department of Informatics and Telecommunications, National and Kapodistrian University of Athens, Athens 15784, Greece.
    \IEEEcompsocthanksitem All authors are with Cogitat Ltd., London, U.K.
    }%
}

\IEEEtitleabstractindextext{%
\begin{abstract}
    \textit{Objective}: Deep learning approaches to the decoding of brain activity from electroencephalography (EEG) signals have gained traction in recent years, with many claims to state-of-the-art classification accuracy. However, despite the convincing achievements on generally accepted benchmarks, the successful translation to clinical or consumer applications appears limited. The frequent disconnect between deep learning performance on controlled brain-computer interfacing (BCI) benchmarks and its lack of generalisation to practical settings indicates hidden overfitting problems that go beyond the need for improving metrics on supervised learning tasks.
    \textit{Approach}: We introduce Disentangled Decoding Decomposition (D\textsuperscript{3}), a weakly supervised method for training deep learning models across heterogeneous EEG datasets. By predicting the place in the respective trial sequence from which the input window was sampled, EEG-D\textsuperscript{3} separates latent components of brain activity, akin to non-linear independent component analysis (ICA). We utilise a novel model architecture with fully independent sub-networks for strict interpretability. To advance our understanding of the underlying signal artefacts and neural processes, we outline a feature interpretation paradigm to contrast the component activation profiles on different datasets and inspect the associated temporal and spatial filters. We then select a subset of latent components for use in downstream applications to avoid overfitting on spurious features caused by task-correlated artefacts.
    \textit{Main Results}: The proposed method reliably separates latent components of brain activity on motor imagery data, including event-related synchronisation (ERS), event-related desynchronisation (ERD), event-related potentials (ERPs), and eye-blink artefacts. Training downstream classifiers on an appropriate subset of these components prevents hidden overfitting, which severely affects end-to-end classifiers trained on conventional class labels. We further exploit the linearly separable latent space resulting from the proposed pre-training method for effective few-shot learning on sleep stage classification.
    \textit{Significance}: The ability to separate genuine components of brain activity from spurious features results in deep learning models that avoid the hidden overfitting problem and generalise well to real-world applications, while requiring only minimal labelled data. With interest to the neuroscience community, the proposed method gives researchers a tool to separate individual brain processes and potentially even uncover heretofore unknown dynamics.
\end{abstract}

}

\begin{document}

\maketitle
\IEEEdisplaynontitleabstractindextext

\IEEEpeerreviewmaketitle

\IEEEraisesectionheading{\section{Introduction}}
\label{sec:introduction}

\IEEEPARstart{T}{he} simultaneous firing of large populations of neurons in response to external stimuli and internal processes results in complex patterns of brainwave activity. These patterns can be observed with electroencephalography (EEG) recordings and studied to gain neuroscientific insights into the workings of the human brain. EEG signals can further be decoded for various practical applications.

Brain-computer interfaces (BCI), for instance, translate brain activity related to movement intentions into digital commands, helping patients with neuromuscular impairments regain basic functional control \cite{wolpaw2002brain}. Popular BCI paradigms are motor imagery (MI) and motor execution (ME), both of which exhibit similar brain activity \cite{abiri2019comprehensive}. Both MI and ME can be decoded from EEG signals, localised over the motor cortex, by extracting their characteristic event-related desynchronisation (ERD) of the mu rhythm during a motor task, which is followed by event-related synchronisation (ERS) upon relaxation \cite{pfurtscheller1999event}.

A routine application of the analysis of EEG signals is in sleep stage classification, as the different sleep stages exhibit markedly specific patterns in the EEG signal that can be observed by a trained clinician \cite{silber2007visual}. The sleep stages are broken down into an awake phase, rapid eye-movement (REM) sleep, and three stages of non-REM sleep named N1, N2 and N3. The analysis of sleep recordings is beneficial in various contexts, including healthcare and consumer technology, but requires a trained clinician for interpretation or large amounts of labelled data to train automated classifiers.


While decoding of EEG brain activity has many applications, it frequently suffers from the presence of signal artefacts, such as ocular activity. Not only do such artefacts reduce the signal-to-noise ratio, but they can actually be class-discriminative, leading to a hidden overfitting problem on the training dataset \cite{frolich2015investigating}. This problem can be partially mitigated by carefully applying artefact removal techniques, such as independent component analysis or linear regression \cite{uriguen2015eeg, jiang2019removal}. However, most deep learning literature on EEG decoding does not apply or mention artefact removal in their methods \cite{craik2019deep}, which likely inflates reported classification results, also depending on the choice of input window extracted from the trial \cite{bakas2023latent}.

Aside from signal artefacts, which can feasibly be removed with appropriate methods, true brain activity can also have similar effects. One example of such activity are residual eye blink-related potentials, resembling event-related potentials (ERPs), that are evoked by the off-on light stimulus and appear to remain in occipital areas even after applying artefact removal \cite{berg1988eyeblink}. Another case are visual evoked potentials (VEPs) elicited by the task-specific cue presentation, which can be used for classification of that task \cite{trocellier2024visual}. Furthermore, posterior alpha power can be selectively suppressed by internal attention paid to the spatial location of items that were presented to be memorised, even after the cue disappears from the screen \cite{pavlov2022oscillatory}. This effect could potentially occur with memorisation of the cue arrows used in a standard motor BCI paradigm, leading to class-discriminative brain activity independent of the actual motor task. While any number of such brain activity phenomena can increase classification performance on the training data, which is usually obtained in a highly controlled laboratory environment, such classifiers are unlikely to transfer well to real-world settings and self-paced movements.

Rather than attempting to remove all of such known and unknown class-discriminative components from the raw signal manually, we propose to leverage the flexibility of deep representation learning to automatically disentangle the signal into latent components. These will include both the genuine components related to the intended task, as well as other class-discriminative activity. We can then select from the set of trained latent components only those that we wish to employ in our classifier.

We therefore propose an interpretable architecture consisting of independent sub-networks, that each are associated with a unique spatio-temporal pattern that reflects distinct brain dynamics. This architecture will be trained in a weakly supervised setting to identify and map the location of the input window within the sequence of the full trial context, utilising the consistent patterns of brain activity that characterise different behavioural tasks. Training across heterogeneous datasets with different experimental designs is achieved by using a shared feature extraction stage, with individual sequence mappings unique to each dataset. This approach allows the scaling-up of model pre-training to aid the discovery of disentangled and generalisable latent representations.

In summary, the main contributions of this paper are as follows:
\begin{itemize}
    \item We present an interpretable architecture with fully independent sub-networks
    \item We devise a weakly supervised representation learning method for disentanglement of latent brain dynamics
    \item We derive latent components of brain activity for motor imagery BCIs
    \item We demonstrate effective few-shot learning on sleep stage classification.
\end{itemize}

We will introduce the method on the example of motor imagery BCIs, and then further apply it on sleep stage classification to show its general applicability. Section \ref{sec:related_work} will summarise related work, followed in section \ref{sec:methods} by the elaboration of the methods as exemplified on the motor task. The main results will be presented in section \ref{sec:main_results}, and few-shot learning on sleep stage classification will be demonstrated in section \ref{sec:other_tasks}. Finally, section \ref{sec:discussion} and section \ref{sec:conclusion} will offer a discussion and conclusion, respectively.

\section{Related Work}
\label{sec:related_work}


Our aim of separating the underlying components of brain activity follows in the spirit of independent component analysis (ICA) \cite{bell1995information}, which linearly separates a multichannel signal into maximally independent components. The response consistency of different independent components (ICs) across trials of the same task has been quantified as inter-trial coherence (ITC) and used as an analysis tool to identify motor components inherent to EEG activity \cite{melnik2017eeg}. ICA is generally used for removal of signal artefacts, like ocular activity \cite{uriguen2015eeg}. However, this approach has numerous limitations. First, it is generally not performed when training deep learning models \cite{craik2019deep}, and it is unclear to what degree ICA cleaning would prevent a strong deep learning model from recovering artefact remnants remaining within the signal. Further, we cannot expect ICA cleaning to remove true components of brain activity that are nevertheless spuriously correlated with our task of interest, such as ERP responses on a motor imagery task. Lastly, even between the components that are relevant to the task of interest, such as the ERD and ERS, there is substantial utility in separating them for targeted application in downstream tasks.

Unlike ICA-based research, which relies on linear algebra, our aim is to find a non-linear transformation of the EEG signal into its constituent components. Hyvarinen et al. \cite{hyvarinen2023nonlinear} argue that non-linear independent component analysis is a challenging proposition due to the non-identifiability of the transformation, i.e. the lack of uniqueness. However, they propose that identifiability can be achieved by providing additional information to the model. One such approach is to assume that true latent components exhibit a degree of auto-correlation. Our approach is based on the same assumption of auto-correlation, since we ask the model to classify nearby input windows as the same label, and furthermore, across trials the component needs to have a consistent response on average. Another assumption they propose is that true latent components exhibit non-stationarity, which would allow for time-contrastive learning (TCL) by identifying the time segment from which an input window was sampled. We apply this in our work as the sequence classification task performed across trials of each dataset.

A popular instantiation of non-linear ICA is contrastive learning, which has recently been done on EEG signals by Banville et al. \cite{banville2021uncovering}, who applied three contrastive learning variants on sleep staging and pathology detection to find a suitable representation space. For the first approach, they sampled pairs of input windows from a continuous EEG signal and depending on the temporal distance between the windows, the model had to classify them as being the same or a different label. The task in the second approach was to classify whether the three provided input windows were in the correct sequence order. In the third approach, the model is asked to identify which of multiple provided samples chronologically followed a set of anchor samples. The authors report that the trained models exhibited a clearly structured latent space, resembling the relevant dimensions of interest for the respective training data, as well as improved downstream classification performance in the low data regime.

Our proposed training task is related to such contrastive learning methods, in that different input windows sampled closely together should be classified as the same and input windows sampled from further away should be classified differently. However, instead of assuming that training data is fully unlabelled, we make use of the ubiquitous trial-based structure available in the BCI field. We recognise that not only are input windows that are sampled closely together likely representing the same brain activity, but also windows sampled from the same time point across trials likely share the same features. An example would be the ERP response, which reliably follows the cue stimulus across trials. While we assume that the basic trial timing of our training data is known and in that sense labelled, it must be noted that individual components of brain activity are nevertheless unlabelled. Unlike in supervised learning, rather than predicting an explicit label, we ask the model to separate the different windows sampled from the same trial.

Another related method are beta-variational autoencoders (beta-VAE), which are trained by reconstructing the input signal after passing it through a regularised bottleneck that enforces uncorrelated latent components \cite{higgins2017beta}. Such models have been applied on EEG data for clinical tasks \cite{honke2020representation} and on emotion recognition \cite{hagad2021learning}. However, beta-VAEs have a number of limitations that arguably make them suboptimal for learning disentangled representations on EEG data. These include the difficulty in designing a loss function that appropriately guides the reconstruction of EEG signals (see \cite{barmpas2025advancing}), as well as the additional complexity associated with training a decoder network that is then fully discarded for classification. Most importantly, we argue that the true latent components inherent in EEG activity are not necessarily uncorrelated, especially given the often highly structured nature of available datasets, and neither can their activity be assumed to be normally distributed. Furthermore, existing beta-VAE architectures generally do not lend themselves to direct interpretation of the discovered components via temporal and spatial filter analysis.

Our approach of using a shared feature extraction stage with multiple classification heads for different datasets was introduced by \cite{huang2013cross} to facilitate cross-language knowledge transfer. The model was thereby trained to recognise multiple languages, improving accuracy when compared to language-specific models. This architecture goes back to \cite{caruana1993multitask} and has recently been used by multiple winning teams in the NeurIPS Benchmarks for EEG Transfer Learning competition to train a shared model across datasets \cite{wei20222021}. We make use of this approach here to train a shared set of latent components across heterogeneous datasets.


\begin{figure*}[htp]
    \centering
    \includegraphics[width=\textwidth]{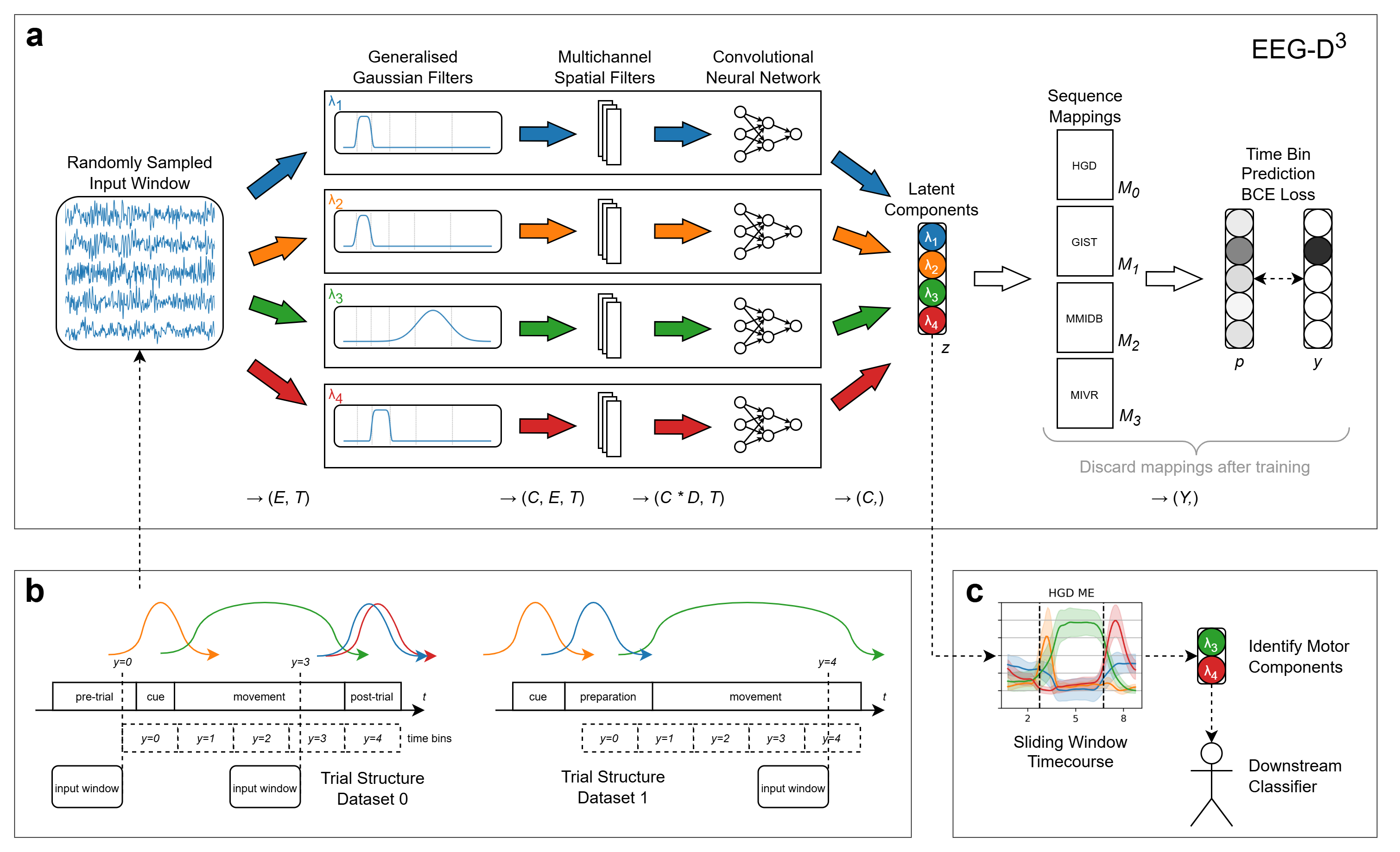}
    \caption{Diagram of the proposed disentangled decoding decomposition method. a) The model architecture consists of $C$ independent sub-networks, implemented using grouped convolutions, each resulting in a single scalar latent component (limited to 4 for illustration purposes). The input signal with electrodes $E$ and window length $T$ is processed by a single Gaussian temporal filter per component, followed by a set of $D$ linear spatial filters. Multiple layers of depthwise separable grouped convolutions and non-linearities are applied for higher-level feature extraction, before reducing the feature dimension down to a single channel per component. A global average pooling over the time dimension is then applied (not shown), resulting in the final latent component vector $z$, which consists of a single scalar for each latent component. This represents a strong bottleneck, forcing reuse of the same components across datasets, which leads to improved separation and generalisation. A trainable linear sequence mapping $M_i$ is indexed, depending on the dataset $i$ associated with the current trial, to make a prediction $p$ that matches the one-hot encoded target vector $y$. b) The training task consists of predicting the time point in the trial sequence $t$ from which the window was sampled out of $Y$ time bins (limited to 5 bins for illustration purposes). The diagram shows two exemplary datasets with several randomly sampled input windows, each with their associated classification targets $y$. The bins are equally spaced, irrespective of the trial structure, as we avoid making assumptions about the brain activity within the trial. c) After discarding the sequence mappings, the activity of the latent components is plotted in a sliding window fashion to produce timecourse plots over known trials, which aid in identifying motor components. This subset of clean components can then be used with downstream classifiers.}
    \label{fig:architecture}
\end{figure*}

\section{Methods}
\label{sec:methods}

The following section describes Disentangled Decoding Decomposition (EEG-D\textsuperscript{3}), a model architecture and associated weakly supervised training task for separating latent components of brain activity. We further describe our approach to feature interpretation in the trained model, as well as the use of selected features in downstream classification tasks. Details for the datasets and their preprocessing are given, as well as for the hyperparameters used during training.

\subsection{Model Architecture}
\label{sec:architecture}

\begin{table*}[!t]
\renewcommand{\arraystretch}{1.5}
\centering
\caption{Table of the different modelling layers comprising the proposed architecture. An input signal with $E$ electrodes and $T$ time points is processed into $C$ latent components. Each latent component consists of its own sub-network, implemented using grouped convolutions. Each component sub-network applies $D$ spatial filters, followed by separable convolutions with $F_1$ and later $F_2$ filters. The resulting latent components are used by the sequence mapping specific to each dataset to predict the bin among $Y$ time bins from which the sample was drawn.}
\label{tab:architecture}
\begin{tabular}{@{}llllll@{}}
\hline
Layer                      & \# Filters  & Kernel Size  & Output Shape                        & Groups   & Options        \\ \hline
Input                      &             &              & ($E$, $T$)                          &          &                \\
GeneralisedGaussianFilter  & $C$         &              & ($C$, $E$, $T$)                     &          & using FFT      \\
DepthwiseConv2d            & $D$         & ($E$, 1)     & ($C$ * $D$, $T$)                    & $C$      &                \\
BatchNorm1d                &             &              & ($C$ * $D$, $T$)                    &          &                \\
Leaky ReLU                 &             &              & ($C$ * $D$, $T$)                    &          &                \\
SeparableConv1d            & $F_1$       & 81           & ($C$ * $D$ * $F_1$, $T$)            & $C$      & padding=40     \\
BatchNorm1d                &             &              & ($C$ * $D$ * $F_1$, $T$)            &          &                \\
Leaky ReLU                 &             &              & ($C$ * $D$ * $F_1$, $T$)            &          &                \\
AveragePool1d              &             & 4            & ($C$ * $D$ * $F_1$, $T$//4)         &          &                \\
Dropout                    &             &              & ($C$ * $D$ * $F_1$, $T$//4)         &          & p=0.25         \\
SeparableConv1d            & $F_2$       & 21           & ($C$ * $D$ * $F_1$ * $F_2$, $T$//4) & $C$      & padding=10     \\
BatchNorm1d                &             &              & ($C$ * $D$ * $F_1$ * $F_2$, $T$//4) &          &                \\
Leaky ReLU                 &             &              & ($C$ * $D$ * $F_1$ * $F_2$, $T$//4) &          &                \\
GlobalAveragePool          &             &              & ($C$ * $D$ * $F_1$ * $F_2$)         &          &                \\
Dropout                    &             &              & ($C$ * $D$ * $F_1$ * $F_2$)         &          & p=0.25         \\
PointwiseConv1d            &             & 1            & ($C$,)                              & C        &                \\
Sigmoid                    &             &              & ($C$,)                              &          &                \\
SequenceMappings           &             &              & ($Y$,)                              &          & indexed by dataset  \\ \hline
\end{tabular}
\end{table*}

The proposed architecture makes use of trainable Gaussian temporal filters as the first layer \cite{ludwig2024eegminer}. These filters are parameterised by a generalise Gaussian function over frequencies, which is initialised to have Gaussian shape and can be trained towards a sinc-like box filter. The initial Gaussian shape has wide transition bands, which results in better gradient support compared to the very narrow transition bands of a sinc filter. The filters are applied after fast Fourier transform (FFT), before going back to the time domain. The generalised Gaussian filter $F$ is defined as

\begin{equation}
\label{eq:generalised_gaussian_filter}
\begin{split}
    F(x) = & e^{-(|x-\mu|/\alpha)^\beta} \\
    \alpha = & \frac{h}{2\ln(2)^{1/\beta}}
\end{split}
\end{equation}

over frequencies $x$, parameterised by the centre frequency $\mu$, bandwidth $h$ and shape $\beta$.

The Gaussian temporal filters are followed by a set of linear spatial filters (see Figure \ref{fig:architecture}, Table \ref{tab:architecture}), in the same vein as EEGNet \cite{lawhern2018eegnet}. After this temporal-spatial processing, the first BatchNorm \cite{ioffe2015batch} and leaky-ReLU non-linearity \cite{maas2013rectifier} are applied.

The temporal and spatial filter layers are followed by two higher-level feature extraction blocks \cite{chollet2017xception}. These consist of a depthwise separable convolution, BatchNorm, leaky ReLU, temporal pooling and dropout \cite{srivastava2014dropout}. The first block uses average pooling, while the second and last block applies global average pooling over the time dimension \cite{lin2013network}. This is followed by a point-wise convolution that combines all channels of each latent component into a single output. This results in a feature vector of length equal to the number of latent components.

The architecture is effectively split into separate sub-networks equal to the desired number of latent components, defined by a model hyperparameter. This is achieved via grouped convolutions, preventing any information flow between components. For this, the convolutional layers are set up in such a way that any cross-channel weights are applied only within each respective group.
Using grouped convolutions allows us to directly associate each temporal and spatial filter with a specific latent component. In addition to improved interpretability, we expect this to help with the separation of different dynamics.

\subsection{Training Task}
\label{sec:training_task}

During training, we ask the model to classify the time bin of the trial from which the given input window was taken. It is worthy of note that this approach to pre-training does not rely on any explicit class labelling of the training trials. Each trial is assigned a fixed number of time bins that serve as training targets, which in the case of the motor task include the pre- and post-action periods (see Figure \ref{fig:architecture}). The model performs this task by mapping the latent components to the respective trial sequence. For a prototypical motor task, the underlying component sequence could take the form of an ERP after the cue presentation, followed by a reduction in eye blinks while the subject is concentrating and an ERD response caused by the motor action, culminating in an ERS at the offset of the motor action and increased eye blinking post-trial.

\begin{figure*}[!ht]
    \centering
    \includegraphics[width=\textwidth]{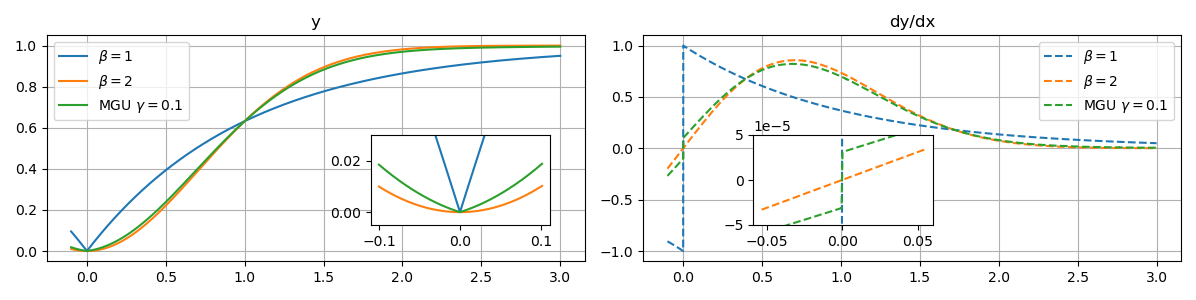}
    \caption{The Mixed Gaussian Unit (MGU) activation function is applied on the sequence mappings to limit the range of their weights to (0, 1). It consists of a Laplacian and a Gaussian function with an interpolation factor $\gamma$. Both functions are special cases of the generalised Gaussian function, controlled by the shape parameter $\beta$. The MGU inherits the stable derivative of the Gaussian function, while preventing diminishing gradients around zero.}
    \label{fig:mgu_activation}
\end{figure*}

Each training dataset is assigned a unique trainable mapping matrix. This is required because different datasets correspond to different sequences of latent components, such as longer or shorter action periods, or different types of cue timings. For each dataset (index $i$), the unique mapping matrix $M_i$ performs a linear weighting of latent components $z$ to obtain the final predictions $p$ as
\begin{equation}
    p=M_iz.
\end{equation}
A binary cross-entropy loss is applied on each output against the one-hot encoded target vector to identify the correct time bin. The sequence mappings are only required for pre-training, and are then discarded for downstream tasks.

To achieve easily readable latent component profiles, we apply a sigmoid activation function on latent components, which limits their output range to (0, 1). These independent outputs are then remixed by the sequence mappings. To constrain the sequence mappings, we apply a custom inverted mixed Gaussian unit (MGU) activation function. The MGU is defined as
\begin{equation}
    MGU(x) = \gamma (-e^{-|x|} + 1) + (1 - \gamma) (-e^{-x^2} + 1),
\end{equation}
where $\gamma$ is the interpolation factor that determines the step size of the derivative around zero and thereby prevents vanishing gradients (see Figure \ref{fig:mgu_activation}). After applying the MGU, we semi-normalise each row vector $m$ of the sequence mappings by dividing it by its sum if the sum is larger than one, to obtain $m^*$ as
\begin{equation}
\begin{split}
    \hat{m} &= MGU(m) \\
    m^* &= \frac{\hat{m}}{ReLU(\sum\hat{m} - 1) + 1}.
\end{split}
\end{equation}
Taken together, this activation is similar to a softmax activation, but prevents the small number of fully flat components we observed when using the softmax function instead. This semi-normalised mapping matrix is then multiplied with the latent components in a linear fashion. Thereby, each final output is a weighted mean of latent components, classifying the sequence bin from which the input window was drawn.

There are multiple incentives embedded in the architecture and training task to separate latent activity into independent components. The first is that the model is required to distinguish different time periods of each trial, which involve distinct patterns of underlying brain activity. As such, if two components are independent in time, the model will separate them. Second, we limit each component to a single Gaussian filter, thereby excluding other frequency bands, and separate components into independent sub-networks, as described above. And third, the limited number of latent components need to be re-used across datasets with different latent structure. As an example, the ERS might coincide with increased eye blinking in one dataset, but not in another dataset, which would lead the model to separate those components.

\subsection{Feature Interpretation}
\label{sec:interpretation}

After training the model to distinguish different parts of the presented motor trials, we can proceed to interpret the resulting latent components. Since the architecture is split into fully separate sub-networks, it is straightforward to attribute the relevant temporal and spatial filters to each component. Specifically, we can plot the magnitude response of the trained Gaussian filters over frequencies to get an indication of the targeted frequency band \cite{ludwig2024eegminer}. The trained linear spatial filters can similarly be interpreted by taking their absolute values and averaging them per component to obtain a single relevance vector over electrodes. This vector can be used to estimate the targeted spatial regions of each component. While this interpretation is not sufficient to specify the exact nature of the trained features, such as band power or synchronisation, it nevertheless offers valuable information to allow us to interpret the different components.

In addition to the temporal and spatial filter interpretation of each component, we can use their response in different task settings to further characterise them. One such test could be the response to eye blink artefacts. For a comprehensive analysis, we run the trained model over a number of different tasks in a sliding-window fashion and record the activity of each latent component over time, dubbed "timecourse" for the purpose of this study. This allows us to both improve our understanding of the nature of the trained components, as well as the nature of the different datasets and the activity they contain. These timecourse plots can further be set next to average plots of the raw signal to gain some insight with respect to the lower frequencies, such as eye blinks or event-related potentials.

To quantify the response of each component in the different settings, we compute the concordance correlation coefficient (CCC) \cite{lawrence1989concordance} between the component responses on different trials of the same subject and setting, and call this metric "timecourse consistency" (TC) from here on. This metric measures how consistently each component reacts in different circumstances, such as a motor execution trial or an isolated eye blink. The TC is given by:
\begin{equation}
\begin{split}
    r_{XY} &= \frac{cov(X, Y)}{\sigma_X \sigma_Y}
        = \frac{\sum_t (x_t - \bar{x}) (y_t - \bar{y})}{\sqrt{\sum_t (x_t - \bar{x})^2} \sqrt{\sum_t (y_t - \bar{y})^2}} \\
    CCC_{XY} &= \frac{2 r_{XY} \sigma_X \sigma_Y} {\sigma_X^2 + \sigma_Y^2 + (\bar{x} - \bar{y})^2} \\
    TC &= \frac{1}{(N^2-N)/2} \sum_{i>0} \sum_{k<i} CCC_{ik}
\end{split}
\end{equation}
For each latent component, we first compute the Pearson correlation $r_{XY}$ of the component response over time $t$ between the timecourses $X$ and $Y$ on a pair of two trials. This correlation is normalised to obtain the CCC. Such pairwise coefficients are computed between trials $i$ and $k$ across all $N$ trials and the lower triangular of the resulting correlation matrix is averaged, excluding the diagonal. This results in the timecourse consistency metric $TC$. A consistency score of 0 means that there is no correlation between the component responses over trials, while a score of 1 means that the component behaves perfectly identically across trials. We choose the CCC over the simple Pearson correlation coefficient, because it accounts for shifts in the prediction distribution across trials, which would negatively impact classification performance.

This component interpretation is repeated separately for each cross-validation fold, as there is no guarantee that the trained components will be in the same order. After assigning interpretations to the relevant components, we can get their mean and standard deviation across folds.

\subsection{Datasets and Preprocessing}
\label{sec:datasets}

The first motor dataset we use is HighGamma (HGD), which includes EEG recordings from 14 subjects performing trials of motor execution by sequential finger tapping \cite{schirrmeister2017deep}. This dataset also includes control trials, which present the same stimulus but without having the subject perform the corresponding action. We use these control trials only for model evaluation. We cut this dataset into trials of 9.5 seconds and re-reference to the P9 electrode.

The next motor dataset is the motor movement/imagery database (MMIDB) from the PhysioNet databank \cite{goldberger2000physiobank, schalk2004bci2000}, which is collected from 109 subjects performing motor execution and imagery trials by repeatedly opening and closing their hands. We cut this dataset into trials of 9.5 seconds and drop subjects 88, 89, 92, 100, 104 and 106 due to recording inconsistencies. Since the P9 electrode is not available on this dataset, we instead re-reference to T9. This dataset is not correctly scaled and we could not find the least significant bit information to perform the rescaling. This is the reason for using signal standardisation for each dataset, as explained below.

We use the Gwangju Institute of Science and Technology motor imagery (GIST MI) dataset \cite{cho2017eeg}. It includes EEG recordings from 52 subjects performing motor execution and motor imagery tasks. Participants were instructed to consecutively touch their thumb with each of the fingers of their hand. This dataset is cut into trials of 7 seconds. Since these trials are shorter than the other datasets used for model training, we temporarily apply zero-padding in the time dimension to be able to stack them together with the other datasets. The input window sampler then ignores the zero-padded period. The 7 second sequence is nevertheless split into the same number of time bins, which therefore have shorter width. This only results in a somewhat higher resolution in the labelling and is not expected to affect training otherwise. We re-reference the dataset to the P9 electrode.

The last motor dataset we include for model training contains EEG recordings from 26 subjects performing a motor imagery task in a virtual reality environment (MIVR) \cite{vagaja2024avatar}. This dataset is cut into trials of 19 seconds, which we split into two halves of 9.5 seconds each. We re-reference to the TP9 electrode.

During dataset preparation we apply Notch filters at 50Hz and 60Hz equally across all datasets. This is followed by a 3rd order zero-phase Butterworth bandpass filter in the 0.05-79.5Hz band. The recordings are then cut into trials centred on the movement task, with some potential overlap on the inter-trial period between trials. This is followed by downsampling to 160Hz with the fast Fourier transform method. For a simple way to combine datasets with different electrode layouts, we use only the 28 electrodes common across all four motor datasets, providing full-head coverage\footnote{Electrodes: Fp1, Fp2, F7, F3, Fz, F4, F8, FC5, FC1, FC2, FC6, T7, T8, C3, Cz, C4, CP5, CP1, CP2, CP6, P7, P3, Pz, P4, P8, O1, Oz, O2}.

During model training, we randomly select windows of 1.5 seconds from the EEG trials and apply a 3rd order zero-phase Butterworth bandpass filter in the $8-40$Hz band. We found the highpass filter to be beneficial for learning motor components more reliably, while still allowing the model to reconstruct some of the lower frequencies. Each electrode within a window is then standardised independently, followed by common average re-referencing (CAR). Performing CAR after standardisation is an unconventional choice that no longer performs strict re-referencing, but has the advantage of reducing the spread of high-power signal artefacts across other electrodes.

\subsection{Setup Details and Hyperparameters}
\label{sec:hyperparameters}

We train the model of our proposed method across all four datasets with $16$ latent components and $16$ time bin targets. At initialisation the Gaussian temporal filters are centred at $24$Hz with a bandwidth of $48$Hz. Given the sampling rate of $160Hz$, the kernel size of the first temporal convolution is set to $81$, with zero-padding of $40$ samples on each end of the signal. The pooling layer has a kernel size of $4$, followed by the second temporal convolution with kernel size $21$ and padding $10$. The model is trained for $100$ epochs using an AdamW optimiser~\cite{loshchilov2017decoupled} with a batch size of $32$, setting the learning rate equal to $0.001$ and weight decay equal to $0.01$.
The datasets are oversampled to target a uniform distribution for equal representation, including the oversampling of executed movement trials in the GIST dataset to match its motor imagery trials. The combined dataset is then split into five subject-independent cross-validation folds.

After the completion of model training and the selection of the two motor components (i.e. ERD and ERS) for a downstream motor task, the mapping matrices are discarded and the rest of the model is kept frozen, serving as a feature extractor for training the linear classifier on top of it. For a given downstream motor task, each network is trained on all four datasets with the same $8-40$Hz bandpass filter for $100$ epochs using an AdamW optimiser with a batch size of $32$, setting the learning rate equal to $0.001$ and weight decay equal to $0.01$. All trained downstream classification models are then separately evaluated against the held-out validation subjects on the pre-trial baseline period and on the control trials of the HGD dataset. The other datasets do not include control trials.

\subsection{Downstream Classification}
\label{sec:classification}

The proposed method produces latent components of brain activity. These need to be selected and classified appropriately to provide a decision function. Here, we manually select only the two components we identified as ERD and ERS, and train a very simple linear model on these frozen components. This binary classification model consists of only six parameters, including the two output biases.

For the first baseline model we choose EEGNet \cite{lawhern2018eegnet} for its popularity and established versatility on different EEG decoding tasks. As a more recent architecture we also include EEGConformer \cite{song2022eeg}, a hybrid model based on convolutions and self-attention.

We classify motor imagery and execution trials against the relaxed state without any movement, as would be required for a "brain click" application \cite{zander2010combining, vansteensel2016fully}. The relaxed state is further specified as either the baseline period before onset of motor activity, or a control trial from the HGD motor execution dataset, which exhibits the same stimulus setting as a movement trial, but instructs the participant to remain still.

Both the linear classifier on top of the selected latent components, as well as the benchmark models, are trained in a supervised way across training subjects and across the four motor datasets. Results are reported on the HGD data, because it is the only one with an explicit control condition. The input window on the action trials is either taken from the action start, specified as the $1.5$ seconds immediately following the cue, or it is taken post action, specified as the $1.5$ seconds immediately following the stimulus offset. With these two selections we are aiming to train an ERD and ERS classifier, respectively. The baseline trials are always taken from the $1.5$ seconds immediately preceding the stimulus onset.

\section{Main Results}
\label{sec:main_results}

\subsection{Reversing Artefact Removal}

\begin{figure}[tbp]
    \centering
    \includegraphics[width=\columnwidth]{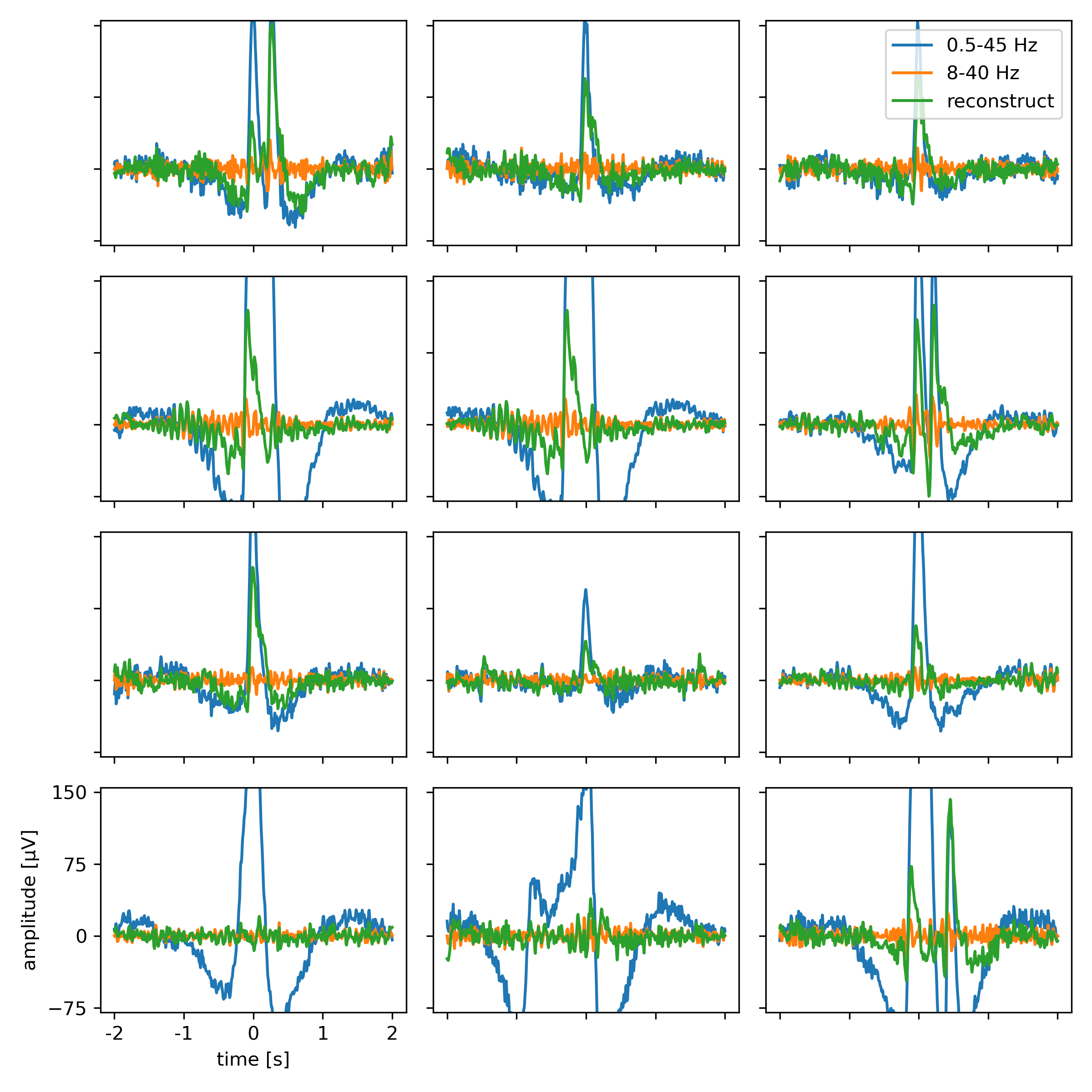}
    \caption{Example eye blink artefacts at the Fpz electrode. A 3rd order Butterworth bandpass filter above 8Hz removes the artefacts visually, but a simple deep learning model (unrelated to our proposed method) can partially undo the bandpass filter to recover most of the eye blinks to use as spurious features during classification. The shown eye blink trials were randomly selected from unseen test subjects on this task, ruling out reconstruction by memorising the trial.}
    \label{fig:eyeblink_reconstruct_examples_val}
\end{figure}

One argument against the necessity of disentangled decoding decomposition is the prospect of careful artefact removal. If all artefacts with spurious task correlations could be removed before model training, this might alleviate the problem of hidden overfitting.

In the following, we will exemplify the risks associated with this approach by attempting to reconstruct eye blink artefacts after applying an aggressive 8-40 Hz bandpass filter that visually removes them completely. We use a simple deep learning model (unrelated to the main architecture proposed above) consisting of two convolutional filter layers with 8 hidden channels, kernel size of 161 and padding 80, separated by a Leaky ReLU activation function. The model was trained on Fpz electrode data from HGD trials for 1000 steps with batch size 32 using an AdamW optimiser with learning rate $0.0003$ and weight decay $0.001$. The training target was to undo the 8-40Hz bandpass filter and reconstruct the 0.5-45Hz signal with mean squared error (MSE) loss. The model input was then the filtered signal and the output was the attempted reconstruction of the original signal. The results on signals centred around eye blinks sampled from unseen test subjects show that a simple deep learning model can recover most of the artefacts from the filtered signal (see Figure \ref{fig:eyeblink_reconstruct_examples_val}).

While other artefact removal techniques might be more effective in removing eye blinks than a bandpass filter, the vast number of possible types of spurious artefacts and components of genuine brain activity make the prospects of signal cleaning daunting, not to mention the widespread lack of any artefact removal strategies employed in EEG deep learning literature discussed earlier. The capacity of deep learning models both enables new possibilities for subject-independent decoding, but it also makes hidden overfitting much more likely to occur.

\subsection{Latent Components}

\begin{figure*}[!ht]
    \centering
    \includegraphics[width=\textwidth]{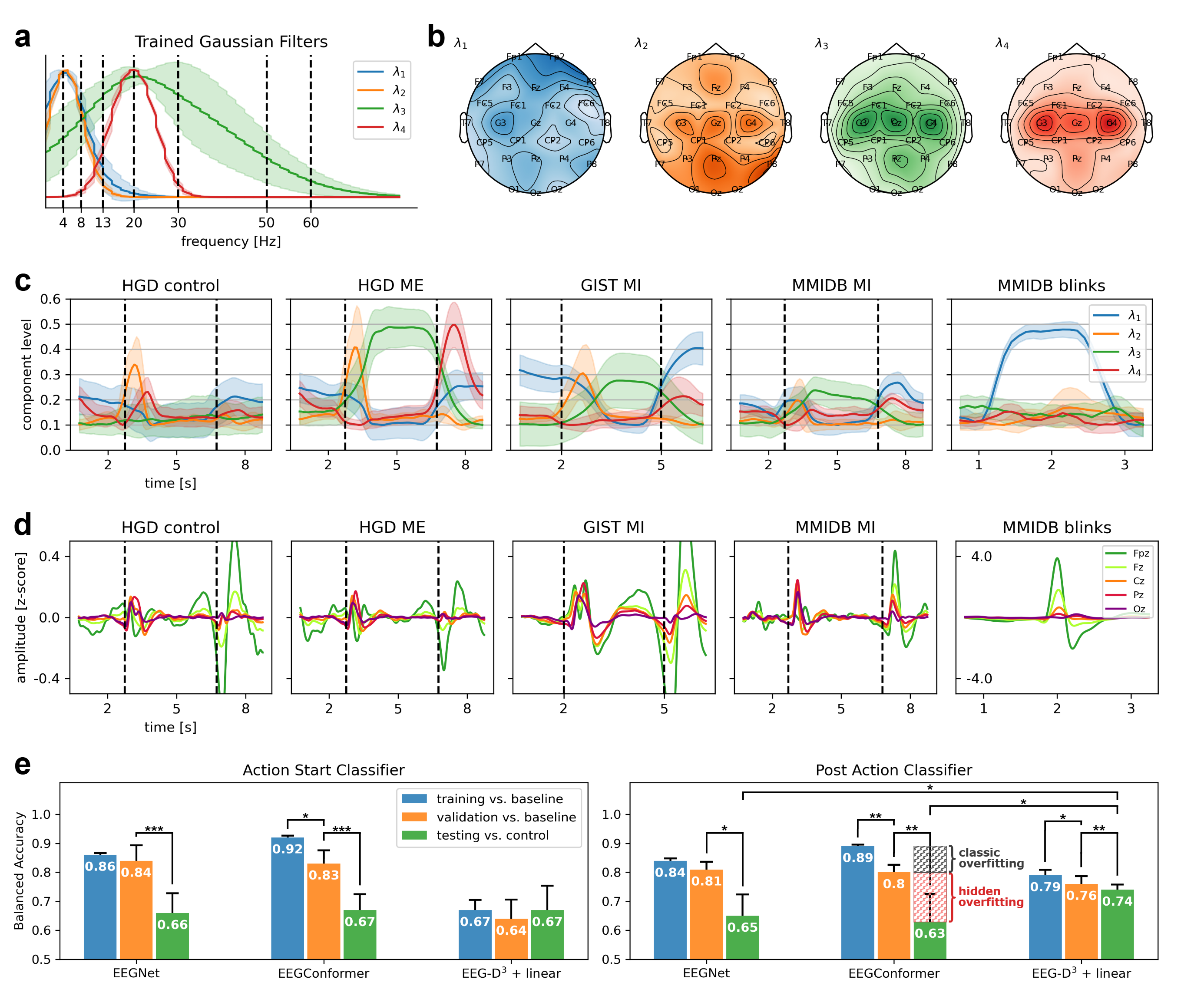}
    \caption{Latent components of brain activity related to the motor paradigm as discovered by our method, and their use in preventing the hidden overfitting problem. Component activity is shown on five different types of EEG trials, including control and left/right hand ME trials from HGD, left/right hand MI trials from GIST and MMIDB, and aligned eye blinks from MMIDB eyes-open resting state. a) Trained Gaussian temporal filters corresponding to the selected components. Shown are the frequency magnitude responses, where the shaded area indicates the standard deviation across folds after manually matching components. b) Trained spatial filters corresponding to the selected components. Shown are the topoplots of the absolute values of the linear spatial filters, averaged across folds. The colour intensity is normalised from zero to the highest value across electrodes. c) Timecourse plots for selected components of the trained model, aggregated over subject-independent fold validation sets. The activation at each time point is plotted at the centre of the model input window. For visual clarity, the mean line of each component is re-centred to align at minimum=0.1 for each subplot, and component $\lambda_1$ is vertically flipped. The shaded area indicates the standard deviation of fold-wise mean lines, thereby corresponding to the reliability of finding the same components across reruns. Dashed lines indicate the start and end of each trial, as marked by the visual stimulus in the respective experimental protocol. We can observe reduced eye blinking activity during the action-phase of each trial (component $\lambda_1$), and aligned eye blinks after the end of the trial. Furthermore, the cue results in an ERP (component $\lambda_2$). The ERD and ERS responses occur during ME and to a lesser extend during MI (components $\lambda_3$ and $\lambda_4$), and neither occur on control and eye blink trials. d) Averaged EEG signals of representative electrodes for each trial type, confirming the presence of ERP responses and timing-synchronised eye blinks. Signals are z-scored with the respective global dataset statistics due to the missing scale information in one of the datasets. Re-referencing is performed as described in section \ref{sec:datasets}. A 3rd order zero-phase Butterworth bandpass filter is applied in range 0.5-4Hz for this plot. To align with the time scale of component timecourse plots, 750ms of the signal are removed from each end. e) Results of benchmark deep learning models on the motor task, highlighting the hidden overfitting problem, and the effectiveness of our proposed method in generalising to unseen test cases. The two tasks tested include classifying the action onset, which includes the ERD motor response, and classifying the post-action period, which includes the ERS response. Balanced accuracies are given on HGD trials as mean (std) over subject-independent cross-validation folds for training and validating on the action window against the baseline preceding the trial, and testing against explicit control trials as an out-of-distribution test case. The strong hidden overfitting of the benchmark models on the dataset structure results in large gaps between the validation and test accuracies. The significance of paired t-tests between training and validation, and validation and control performance are indicated for each model, as well as between the control performance of different models against each other: * $p < 0.05$, ** $p < 0.01$, *** $p < 0.001$.}
    \label{fig:timecourse}
\end{figure*}

\begin{table*}[tbp]
\renewcommand{\arraystretch}{1.5}
\centering
\caption{Timecourse consistency metric for selected components on the different trial types, calculated as the concordance correlation coefficient between the component responses across trials. The eye component has the highest consistency on isolated blink trials, and similar consistency between action and control trials. The ERS and ERD motor components have little to no consistent response on control and isolated blink trials, while exhibiting highest consistency on motor execution trials and modest consistency on motor imagery trials. The results indicate a successful separation of components. The metric is computed per subject-specific recording and then averaged. Results are given as mean (std) over cross-validation folds.}
\label{tab:consistencies}
\begin{tabular}{@{}llllllll@{}}
\hline
                    & HGD control     & HGD ME       & GIST ME      & GIST MI      & MMIDB ME      & MMIDB MI      & MMIDB blinks     \\ \hline
Component 1 (eye)   & 0.10 (0.07)     & 0.15 (0.09)  & 0.13 (0.04)  & 0.25 (0.05)  & 0.10 (0.01)   & 0.11 (0.02)   & 0.37 (0.06)      \\
Component 2 (ERP)   & 0.17 (0.04)     & 0.25 (0.07)  & 0.17 (0.07)  & 0.21 (0.04)  & 0.09 (0.01)   & 0.07 (0.01)   & 0.08 (0.05)      \\
Component 3 (ERD)   & 0.03 (0.02)     & 0.39 (0.03)  & 0.26 (0.07)  & 0.16 (0.05)  & 0.31 (0.05)   & 0.09 (0.03)   & 0.00 (0.01)      \\
Component 4 (ERS)   & 0.05 (0.01)     & 0.30 (0.02)  & 0.19 (0.06)  & 0.10 (0.03)  & 0.23 (0.03)   & 0.05 (0.01)   & 0.00 (0.01)      \\ \hline
\end{tabular}
\end{table*}

After fitting the EEG-D$^3$ model using the proposed weakly supervised training task, the behaviour of each latent component can be inspected to see how it relates to various brain dynamics. Considering the motor task, we identified four latent components of particular interest, enumerated as $\lambda_{1-4}$. For this analysis, the centre frequency and bandwidth of Gaussian filters will be given as the mean (std) over cross-validation folds. Important electrodes are identified with a one-sample t-test with alternative hypothesis that the expected value across folds of the absolute spatial filter weights for the given electrode is greater than the population mean of the filter weights. Provided are the electrodes with $p<0.01$. The component consistency on different datasets is quantified using the concordance correlation coefficient between trials, given as mean (std) over cross-validation folds.

The $\lambda_1$ component uses a Gaussian filter with centre frequency of 3.79Hz (std=1.37Hz) and bandwidth 12.8Hz (std=4.68Hz). The spatial filter weights focus on Fp2 ($p=0.00019$), F8 ($p=0.0041$) and C3 ($p=0.0075$). This component has the highest consistency of 0.37 (std=0.06) on isolated blink trials. Responses on other datasets indicate activity during the inter-trial period, but not during the trial, with consistencies ranging from 0.10 (std=0.07) on HGD control, to 0.25 (std=0.05) on GIST MI. This is consistent with the commonly used instruction to abstain from blinking during the trial. Taking the frequency and spatial filter insights together with the component responses on various datasets, we can identify $\lambda_1$ as an ocular activity component.

Component $\lambda_2$ is centred at 4.49Hz (std=0.35Hz) with bandwidth 9.47Hz (std=1.01Hz). The significant electrodes are P8 ($p=0.00065$), Oz ($p=0.00080$) and Pz ($p=0.0071$). This component shows a single peak of activity at the beginning of the trial for each dataset, including HGD control trials. Consistencies range from 0.07 (std=0.01) on MMIDB MI, to 0.25 (std=0.07) on HGD ME. There is some consistency of 0.08 (std=0.05) on isolated eye blinks. Taken together, we can identify this component as ERP activity with minor reactivity to eye blinks.

Much higher frequencies are seen in component $\lambda_3$, with a centre frequency of 23.04Hz (std=5.29Hz) and bandwidth 38.45 Hz (std=11.19Hz). Using the bandwidth definition of full-width at half maximum, this filter is not frequency selective. Considering the Gaussian shape however, it is still important to note its centre frequency. The significant spatial filter weights are on C4 ($p=0.00010$), C3 ($p=0.00084$), FC1 ($p=0.0011$), Cz ($p=0.0018$) and CP6 ($p=0.010$). Component $\lambda_3$ has no consistency on eye blinks with mean 0.00 (std=0.01) and very low consistency of 0.03 (std=0.02) on HGD control trials. The response is very high during the action phase of motor execution trials, with consistencies of 0.39 (std=0.03) on HGD ME, 0.26 (std=0.07) on GIST ME, and 0.31 (std=0.05) on MMIDB ME. Consistencies reach moderate levels on MI trials, with 0.16 (std=0.05) on GIST MI and 0.09 (std=0.03) on MMIDB MI. Considering these insights, this component can be identified as the ERD response.


Lastly, component $\lambda_4$ is characterised by a centre frequency of 19.78Hz (std=0.29Hz) and bandwidth 11.86Hz (std=0.71Hz). The filter parameters of this component are remarkably stable over cross-validation folds, with a very low standard deviation. The significant electrodes are C4 ($p=0.00012$), C3 ($p=0.00013$), Cz ($p=0.00023$), FC2 ($p=0.0014$) and FC1 ($p=0.0026$). The component shows a single peak after the action phase of each trial, with consistencies having a similar trend to $\lambda_3$, showing no response on eye blinks with 0.00 (std=0.01), and very low response of 0.05 (std=0.01) on HGD control trials. Likewise, consistencies on ME trials are high, with 0.30 (std=0.02) on HGD ME, 0.19 (std=0.06) on GIST ME, and 0.23 (0.03) on MMIDB ME. Regarding MI trials, the component consistencies reach 0.10 (std=0.03) on GIST MI and 0.05 (0.01) on MMIDB MI. Component $\lambda_4$ can therefore be identified as the ERS response following movement offset.

\subsection{Downstream Motor Classification}

In the following, we compare the classification performance of the benchmark models EEGNet and EEGConformer on the action start and post-action windows to the classification performance of a simple linear classifier trained on top of the frozen motor components obtained with our approach.

The benchmark models reach higher classification accuracies on the training task, consisting of classifying the onset of motor activity, or the post action period, against the pre-trial baseline (see Fig. \ref{fig:timecourse}). This result is expected, because the benchmark models are trained on the raw EEG signal and can therefore overfit on hidden task-discriminative, but motor-unrelated components, which we have manually excluded from the downstream classifiers in our approach by removing associated latent components. When transferring these trained benchmark models to perform inference against a different inactive period, however, which in this case consists of an explicit no-action control trial, their classification performance drops greatly. Both architectures lose 18\% and 16\% accuracy on classifying the action start, and 16\% and 17\% on classifying the post-action period. In contrast, our approach performs well on the out-of-distribution control trials, with an increase of 3\% on the action start and a decrease of only 2\% on the post action window, respectively. This demonstrates the practical benefit of prioritising robust decoding models.

\section{Few-Shot Learning on Sleep Data}
\label{sec:other_tasks}

\begin{figure*}[!ht]
    \centering
    \includegraphics[width=\textwidth]{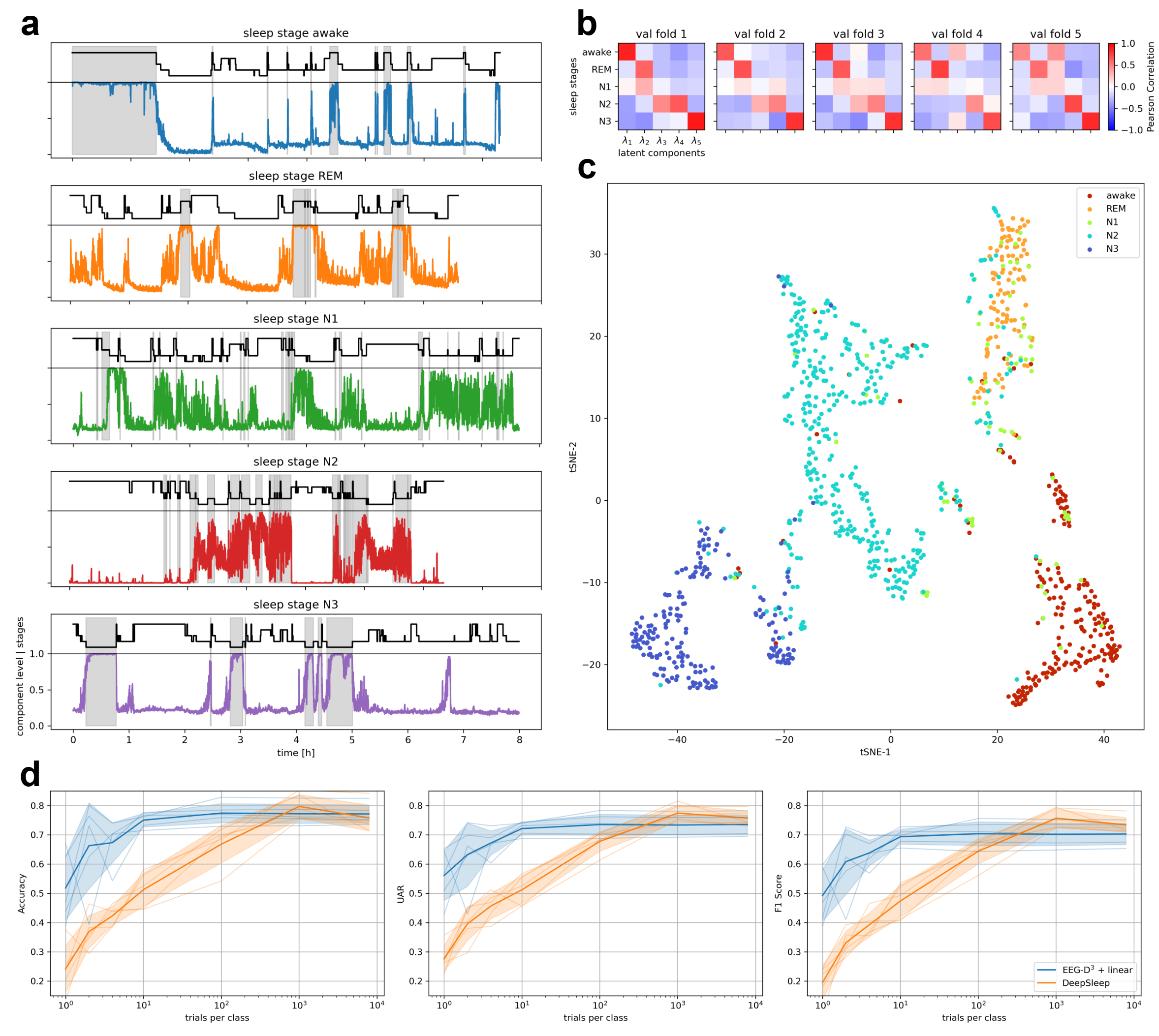}
    \caption{Latent components of brain activity related to the sleep stage paradigm as discovered by our method and their application for few-shot learning. All components are shown after pre-training, without fine-tuning with class labels. a) Examples of component activity levels over recording nights that were randomly chosen from the validation set of the first fold, highlighting a different sleep stage on each subject. The coloured lines show the component that most closely corresponds to each respective sleep stage. The timing of the sleep stage is indicated by the grey shaded area. The stepped line shown above each plot traces the sleep stages as Awake, REM, N1, N2, and N3. All components show low activity over long periods of the sleep recording, with a strong tendency to increase during occurrences of a specific sleep stage. b) Pearson correlation matrices showing the degree to which each latent component tracks a specific sleep stage, for each of the five cross-validation folds. The components were re-ordered manually for readability. The discovered components are generally very consistent and well disentangled across validation folds. c) Dimensionality reduction plot of the latent space components discovered by our method. Shown is a two-dimensional tSNE (t-distributed stochastic neighbour embedding) obtained on 1000 trials randomly sampled across subjects from the validation set of the first cross-validation fold. The sleep stages are well separated, with exception of the difficult to classify N1 stage. d) Few-shot learning of a linear classification head applied on the latent components, compared to a DeepSleep model trained end-to-end on the same number of labelled trials. No fine-tuning of the discovered latent components was performed, as they were frozen during training of the downstream classifier. Results show accuracy, UAR and F1 score aggregated over five cross-validation folds. Our method reaches high levels of classification performance with very few labelled trials per class.}
    \label{fig:sleep_components}
\end{figure*}

The proposed method of weakly supervised representation learning is ideally suitable for separating latent components of brain activity and removing spurious features to improve generalisation. Another benefit of this approach is that the discovered latent space is linearly separable, and therefore presents the possibility of few-shot learning on a small number of labelled examples.

We explore this opportunity on a sleep stage classification task, where we do not expect spurious components that would need to be removed from the latent space. Rather, we will investigate whether the latent components can be directly attributed to specific sleep stages and used for few-shot learning with a linear classifier.

\subsection{Pre-Training Setup}

For the sleep stage classification task we make use of the ANPHY-Sleep dataset \cite{wei2024anphy}, which includes whole night recordings from 29 healthy subjects with high-density EEG. We drop a number of peripheral and non-standard electrodes\footnote{Dropped electrodes: SO1, SO2, F9, F10, ZY1, ZY2, T9, T10, F11,
FT11, FT9, TP11, TP9, P11, F12, FT10, FT12, TP10, TP12, P12, CP5}, resulting in 60 channels. For better alignment across subjects, we clip recordings to a maximum length of 8 hours around the respective mid-point. Preprocessing includes a 3rd-order Butterworth bandpass filter in $0.5-45Hz$, resampling to $200Hz$, and common average re-referencing.

Pre-training is performed by assigning $32$ time bins for each whole night recording, with one time bin covering about 15 minutes of continuous sleep data. The underlying assumption is that typical sleep stages are longer than this time frame. With a window size of $30$ seconds, this setup results in 30 unique windows per time bin for each recording. The model is then trained to classify each randomly sampled $30$-second window into the respective time bin from which it was drawn, as illustrated in Fig. \ref{fig:architecture}. Each separate recording is treated as a different dataset and assigned its own sequence mapping, since sleep stages are not expected to line up across subjects. In cases where the recording is shorter than 8 hours, the length of each time bin is reduced accordingly to match the available duration. Results are aggregated over five subject-independent cross-validation folds.

The model is set up with $5$ latent components, ideally corresponding to the sleep stages. At initialisation the Gaussian temporal filters are centred at $24$Hz with a bandwidth of $48$Hz. Given the sampling rate of $200Hz$, the kernel size of the first temporal convolution is set to $101$, with padding of $50$. The pooling layer has a kernel size of $16$, followed by the second temporal convolution with kernel size $101$ and padding $50$. The other hyperparameters are kept the same as on the motor task, using an AdamW optimiser with learning rate $0.001$, weight decay $0.01$, dropout $0.25$, batch size $32$, and binary cross-entropy loss with label smoothing $0.1$. We run $40$ epochs over the whole training dataset, accounting for the $960$ windows sampled from each recording.

\subsection{Latent Components}

The results of pre-training our proposed method on sleep recordings are shown in Fig. \ref{fig:sleep_components}. We can observe that each of the latent components conforms well to one of the canonical sleep stages. The only exception is the N1 stage, which generally has low inter-rater agreement, resulting in unreliable class labels that might not be clearly identifiable from EEG data \cite{silber2007visual, perslev2021u}. Matching latent components to sleep stages, we can perform a statistical analysis to obtain Pearson correlations between each sleep stage and the respective most relevant component. Since sleep stage time series exhibit strong auto-correlation, we assess significance using correlations with 1000 random-phase surrogates of the component predictions \cite{ebisuzaki1997method}. Correlations and significance values are reported as the mean over fold validation sets. Component $\lambda_1$ on awake obtains $r=0.67$ ($p<0.001$), $\lambda_2$ on REM obtains $r=0.65$ ($p<0.001$), $\lambda_3$ on N1 obtains $r=0.03$ ($p=0.42$), $\lambda_4$ on N2 obtains $r=0.56$ ($p<0.001$), $\lambda_5$ on N3 obtains $r=0.81$ ($p<0.001$).

\subsection{Downstream Sleep  Classification}

To classify the pre-trained latent components into sleep stages, we apply an affine batchnorm and linear layer on all five discovered latent components, with a total of $40$ trainable parameters. To take advantage of the well-calibrated latent component outputs, we divide the linear layer weights by their L1 norm over the outputs dimension. This drives the layer towards assigning each latent component to only a single class and improves performance on very limited training samples.

As an end-to-end supervised learning benchmark we apply the DeepSleep model with $8$ filter channels, filter kernel size $101$ and max-pooling with kernel size $16$ in both convolution blocks, and added batchnorm layers after each temporal convolution \cite{chambon2018deep}.

The linear downstream classifier and benchmark model are trained on a labelled subset of trials from the training set of each subject-independent cross-validation fold, and evaluated on the respective unseen validation set. We conduct scaling tests with the number of training samples per class ranging from a single trial to the full dataset size. Both downstream sleep stage classifiers are trained for 2000 update steps, independent of the number of unique training samples. We use an AdamW optimiser with batch size $32$, learning rate $0.001$, weight decay $0.01$, dropout $0.25$, and cross-entropy loss.

\begin{table*}[tbp]
\renewcommand{\arraystretch}{1.5}
\centering
\caption{Sleep stage classification results of a linear classifier trained on top of the discovered latent components, without further fine-tuning of the model.  Results are compared to a DeepSleep model trained end-to-end on the same number of trials. Metrics include the mean (std) across folds of the accuracy, UAR and F1 score. The linear classifier achieves high performance on very few trials, reaching a ceiling at 100 trials per class, and remains close to and end-to-end trained benchmark model in the large data regime. *For the dataset-sized test the under-represented sleep stages are oversampled to the number of trials available for the most frequent stage.}
\label{tab:sleep_accuracies}
\begin{tabular}{@{}lllllllll@{}}
\hline
         & Trials per class  & 1             & 2           & 4           & 10          & 100         & 1000        & 7878*        \\ \hline
Accuracy & EEG-D\textsuperscript{3} + linear & 0.52 (0.11) & 0.66 (0.15) & 0.67 (0.07) & 0.75 (0.02) & 0.77 (0.03) & 0.77 (0.03) & 0.77 (0.03) \\
         & DeepSleep         & 0.24 (0.08)   & 0.37 (0.06) & 0.42 (0.03) & 0.51 (0.06) & 0.67 (0.06) & 0.80 (0.04) & 0.76 (0.04) \\ \hline
UAR      & EEG-D\textsuperscript{3} + linear & 0.56 (0.08) & 0.63 (0.11) & 0.67 (0.04) & 0.72 (0.02) & 0.74 (0.04) & 0.73 (0.03) & 0.74 (0.03) \\
         & DeepSleep         & 0.28 (0.04)   & 0.40 (0.06) & 0.46 (0.04) & 0.51 (0.05) & 0.68 (0.03) & 0.77 (0.03) & 0.76 (0.02) \\ \hline
F1 Score & EEG-D\textsuperscript{3} + linear & 0.49 (0.09) & 0.61 (0.11) & 0.64 (0.04) & 0.69 (0.03) & 0.70 (0.04) & 0.70 (0.04) & 0.70 (0.04) \\
         & DeepSleep         & 0.19 (0.06)   & 0.33 (0.04) & 0.39 (0.04) & 0.47 (0.05) & 0.64 (0.04) & 0.76 (0.03) & 0.73 (0.02) \\ \hline
\end{tabular}
\end{table*}

Few-shot learning results of the downstream sleep classification task are given in Fig. \ref{fig:sleep_components} and Table \ref{tab:sleep_accuracies}. We observe that the linear classification head trained on as low as one trial per class reaches competitive performance levels, with a large advantage over the end-to-end trained DeepSleep model that requires orders of magnitude more labelled data.

The small drop in performance for the DeepSleep model when going from 1000 trials per class to the full dataset might be explained by the oversampling of minority classes we perform in the latter case, which is required to balance the class counts. This incidental result suggests that undersampling could lead to better performance on this task than oversampling (see \cite{buda2018systematic} for a more systematic discussion).

\section{Discussion}
\label{sec:discussion}

The representation learning approach introduced here successfully disentangles latent components of brain activity and other artefacts contained in the raw EEG signal. Relying only on the segmentation of each datasets into a repeated trial structure, without requiring explicit class labels, we can learn useful representations of brain activity that directly translate to downstream classification performance and generalisation. It is noteworthy that the disentangled latent components are not necessarily uncorrelated, not least due to negative correlations, such as between mutually exclusive ERD and ERS responses and between sleep stages.

Splitting the model architecture into separate sub-networks improves interpretability of the trained components, although it precludes the sharing of low-level weights and features. This could result in less efficient use of parameters and computational resources. However, it is not necessarily clear that EEG exhibits the same compositional hierarchies of features as images do \cite{lecun2015deep}, which means that this restriction might only have limited downside in the EEG domain.

The proposed model inspection approach, including temporal filters, spatial filters, timecourse plots of component activity, and the metric of timecourse consistency, proves invaluable as an interpretation and evaluation method. It allows us to reliably identify various latent components of interest and systematically study their behaviour under varied conditions. Genuine components of brain activity relating to the phenomenon of interest can thereby be separated from other signal components, such as ocular artefacts or stimulus-induced evoked responses.

Utilising the timecourse plots of component activity over different trials, we observe that eye blinks by the subject cannot be assumed to be randomly distributed across the trial. Rather, there are marked patterns in the expected distribution of eye blinks, with a specific prevalence of blinking immediately after the offset of the action period. Because this period contains the rather important ERS response, this finding cannot be neglected when designing a trial segmentation for training motor decoding models. The significance of this is also observed in the severe overfitting of the benchmark models trained in a supervised way on this post-action window against the pre-trial baseline.


The difference in the consistency of identified motor responses between the different datasets could be explained by the type of movement that the respective subjects were instructed to perform. The highest consistency was observed in HGD trials (sequential finger tapping), followed by MMIDB (repeated fist closing) and GIST (sequential finger touching). The finger tapping condition of HGD might expected to produce the most proprioceptive feedback, which could lead to an increased observable brain response \cite{ramos2012proprioceptive}. While HGD did not include MI trials, GIST reached a higher consistency than MMIDB, which could be explained by the increased complexity of imagining individual finger movements \cite{mizuguchi2017changes}.

A limitation of the proposed method is that it only separates components of brain activity that appear at different time points during the same trial, in at least one of multiple datasets. In the context of this study, this means that the method does not distinguish between left and right hand movements, but rather recognises them as the same motor components. In other words, separating left and right hand movements into different components currently offers no benefit in identifying the trial bin from which the input window was taken. Carefully structuring the dataset to include separate left and right hand movements in the same trial could alleviate this shortcoming.

Making use of the linearly separable latent component space, the downstream classification performance of our proposed method achieves strong accuracy levels in a few-shot learning setting. At 100 trials per class it reaches a performance ceiling on this task, beyond which it does not improve. This could be overcome by fine-tuning the feature extraction stage on the labelled trials, rather than only training a linear classifier on top of the latent components, but this fine-tuning step could open up the possibility for task-correlated artefacts to be re-introduced into the latent components. Exploration of this is left for future work.

Important future work is left to be done on the process of defining the latent components. This includes how the choices of input window size and target bin size influence the scale of latent components discovered. It is imaginable, for example, that the different peaks and troughs of an ERP response could be disentangled into their own separate components, instead of obtaining a single "ERP" component.

\section{Conclusion}
\label{sec:conclusion}

We have shown that training deep learning models in a supervised fashion to classify EEG signals can lead to hidden overfitting on task-correlated artefacts and spurious components of brain activity. A straightforward focus on increasing classification performance, even when using valid subject-independent train-test splits, can therefore not be expected to translate to successful real-world inference. It could simply be the case that a novel architecture is better at utilising such spurious features, rather than extracting better representations of the targeted brain activity.

To address this hidden overfitting problem, we introduced a novel weakly supervised training method that leads to disentangled latent representations of the artefacts and components of brain activity contained within the signal. We further introduced an extensive feature interpretation paradigm, utilising the focus on explicit interpretability of the proposed model architecture, as well as the timecourse plots and consistency metric of each discovered latent component in various task conditions. This interpretability then allowed us to discard latent components that are not related to the targeted brain activity, thereby successfully translating the performance of downstream classifiers to unseen test cases. This weakly supervised learning method further enables pre-training on large unlabelled datasets, followed by effective few-shot learning utilising the linearly separable latent components.

The proposed representation learning approach offers the promise of discovering and specifying novel components of brain activity on various tasks of interest. This alone can be of great neuroscientific benefit to the research community. Furthermore, the improved generalisation, resulting from the ability to limit downstream classifiers to the true components of the targeted brain activity, brings us one step closer to real-world subject-independent EEG decoding models.

\ifCLASSOPTIONcaptionsoff
  \newpage
\fi


\bibliographystyle{IEEEtran}
\bibliography{IEEEabrv, references}

\end{document}